# ForeCal: Random Forest-based Calibration for DNNs

Dhruv Nigam
dhruv.nigam@dream11.com
Dream11
Mumbai, India

## ABSTRACT

Deep neural network(DNN) based classifiers do extremely well in discriminating between observations, resulting in higher ROC AUC and accuracy metrics, but their outputs are often miscalibrated with respect to true event likelihoods. Post-hoc calibration algorithms are often used to calibrate the outputs of these classifiers. Methods like Isotonic regression, Platt scaling, and Temperature scaling have been shown to be effective in some cases but are limited by their parametric assumptions and/or their inability to capture complex non-linear relationships. We propose ForeCal - a novel post-hoc calibration algorithm based on Random forests. ForeCal exploits two unique properties of Random forests: the ability to enforce weak monotonicity and range-preservation. It is more powerful in achieving calibration than current state-of-the-art methods, is non-parametric, and can incorporate exogenous information as features to learn a better calibration function.

Through experiments on 43 diverse datasets from the UCI ML repository, we show that ForeCal outperforms existing methods in terms of Expected Calibration Error(ECE) with minimal impact on the discriminative power of the base DNN as measured by AUC.

## 1 INTRODUCTION

Classification machine learning models, especially DNNs are crucial to decision-making processes across various domains like medical diagnosis[6], financial risk assessment[1], and web search[3]. These models help decision-making by generating estimates of the probability of an event given a set of inputs. In a few of these domains, these output probabilities need only to be discriminative i.e. the output of the model for any given input is used to make relative decisions. For example in an ad-based business, if a model suggests that the likelihood of a user clicking an advert A is higher than that for B, A is shown to the user[3]. The absolute values of the predicted probabilities does not matter here, only their positions relative to each other. In such cases, the model used should have good discriminative power as measured by metrics like accuracy and AUC.

However, in many domains, we need the predicted probabilities to reflect the true likelihood of an event. For instance, in the financial sector, lenders require accurate estimates of the likelihood of a loan-default event to make informed decisions on loan approvals. On e-commerce platforms, understanding the true likelihood of a user churning is crucial for devising effective customer retention strategies. In these cases, the model should be well-calibrated in addition to having good discriminative power. In a well-calibrated model, the output probability $p$ corresponding to an event directly corresponds to the likelihood of the event.

Author's address: Dhruv Nigam, dhruv.nigam@dream11.com, Dream11, Mumbai, India, 400051.

Miscalibration arises when model outputs deviate from true likelihoods. It has been shown that Deep Neural Network based classifiers , which are trained to optimize the Negative Log likelihood (NLL) on the training data, tend to overfit on NLL - achieving high discrimination between inputs but poor calibration[7]. Specifically, they tend to be overconfident in their predictions, skewing the predicted probabilities towards 0 or 1(Figure 1). This deviation impairs the DNN's reliability when it comes to tasks where estimating the true likelihood of an event is crucial.

Many techniques have been proposed in recent years to address this[7, 14–17]. These techniques can be divided into two broad categories. The first category of calibration methods modify the training loss of the base model to improve calibration ab initio. This involves training the DNN with a modified Negative Log likelihood which incorporates a penalty for miscalibration[7]. These approaches do not guarantee that the model will retain its discriminative power while achieving better calibration.

The second category of solutions use post-hoc calibration techniques that transform the predictions generated by an uncalibrated model (hereafter referred to as base model) into calibrated probabilities using a *calibration function*. This calibration function is approximated using another model (hereafter referred to as calibration model) that is generally trained on a held-out calibration dataset. This approach is more flexible and can be applied to any base model, not just DNNs, without changing the training process. Depending on the algorithm used to train the calibration model, these methods can also provide guarantees on preserving the discriminative power of the base model.

We posit that any post-hoc calibration method should have the following desirable properties-

- **Non-Parametric** It should not assume any parametric form for the calibration function. This is because the calibration function can be complex and non-linear, and assuming a parametric form can lead to underfitting.
- **Rank preserving** It should preserve the relative ranking of the uncalibrated probabilities and consequently ROC curve and AUC. This ensures that the discriminative power of the base model is not compromised.

This paper presents a novel post-hoc calibration approach - **Fore-Cal** that uses *monotonically constrained Random Forest regression* to learn a post hoc calibration function that improves calibration while preserving discriminative power. The monotonicity constraint ensures that calibrated probabilities are non-decreasing with respect to uncalibrated probabilities.

Random forests[4] is a powerful non-parametric learning technique that achieves close to state-of-the-art performance on tabular data[5]. It also allows for constraints to be imposed on the calibration function, such as weak monotonicity which is crucial to satisfy the rank preservation property. An added advantage, beyond satisfying the above two desirable properties, is that it can incorporate information that can help learn a better calibration function by adding extra *features* apart from uncalibrated probabilities from the base model. These features might not have been available or considered when training the base model or might have been dropped during the training process because of their inability to contribute to the discriminative power of the model.

## 2 RELATED WORK

The field of machine learning has seen a surge in the development of calibration methods in recent years primarily due to the increasing use of machine learning models - particularly DNN-based classifiers - in high-stakes decision-making processes.

The first category of calibration methods is those that modify the training loss of the base model to improve calibration ab initio. Among these, a common approach is adding a differentiable auxiliary surrogate loss for expected calibration error to the Negative Log Likelihood (NLL) loss. Examples of such auxiliary loss functions include Maximum mean calibration error(MMCE)[11], AvUC[9], and Soft Calibration objective[8]. More radical approaches that do away with NLL altogether have also been proposed. Among them is Focal loss[14] which modulates the NLL loss to down-weight observations where the classifier is already confident - limiting the scope for overconfidence and hence, miscalibration. Dual focal loss[18] is an extension of focal loss that improves on the original focal loss, which is overly under-confident in some cases, by modifying the loss function to be more confident.

Recent work has shown that ab initio calibration using modified loss functions and regularisation techniques often does more harm than good[19]. Models trained in such a way are often less discriminative and are less calibratable by post-hoc calibration methods.

The post-hoc calibration approach is thus preferred because of its generality, flexibility, and the fact that it frees the designer of the machine learning model from the need to add additional calibration measures into the objective function used to learn the base model. Under the post-hoc calibration framework, the base model can be trained using any machine learning algorithm that maximizes discriminative power. The calibration model then, can be trained independently using a different machine learning algorithm that maximizes calibration while preserving the discriminative power of the base model.

Examples of post-hoc calibration models include Platt scaling[17], which is a logistic regression based approach. It learns parameters to perform a linear transformation on logits predicted by the base model to obtain calibrated probabilities. Temperature scaling[7] is a single-parameter variant of Platt scaling that learns a single temperature parameter to scale the logits predicted by the base model. Isotonic Regression[21] is a non-parametric approach that fits a piecewise constant, non-decreasing function to transform uncalibrated base model probabilities into calibrated ones. Histogram Binning[20] obtains calibrated probabilities by partitioning the uncalibrated predicted probabilities into bins and using the empirical probabilities in those bins as an estimate of the true likelihood.

Bayesian Binning into Quantiles (BBQ)[15] is an extension of histogram binning uses a Bayesian approach to determine the binning strategy based on the available training data.

Parametric methods like Platt scaling and Temperature scaling are limited by their assumption of a parametric form for the calibration function. Specifically, they assume that the calibration function that transforms uncalibrated logits to calibrated logits is linear. Binning-based methods like Histogram binning and BBQ are non-parametric but are limited by their inability to capture complex non-linear relationships between the input and output. They are also sample inefficient and may lead to inaccuracies when the number of bins is large or there are few observations per bin[10]. They are also piece-wise constant within the bin boundaries and do not provide a smooth calibration function.

## 3 DEFINITIONS

### 3.1 Calibrated classifier

Consider a binary classification task with a dataset $\mathcal{D}$ comprising $\mathcal{X} = \{X_1, X_2, \ldots, X_n\}$, where each $X_i$ is a vector of input features, and corresponding binary labels $\mathcal{Y} = \{y_1, y_2, \ldots, y_n\}$, where $y_i \in \{0, 1\}$.

A binary classifier $g : \mathcal{X} \to [0, 1]$ is a function that assigns a probability estimate $g(X_i)$ to each input feature vector $X_i \in \mathcal{X}$. This represents the classifier's estimated probability that the label $y_i$ corresponding to $X_i$ is 1.

Among the samples $\mathcal{X}_p = \{X_i \mid g(X_i) = p\}$, for whom the classifier $g$ predicts $y = 1$ with probability $g(X_i) = p$, the proportion of samples for which $y_i = 1$ should be approximately $p$ for an ideal classifier.

$$\frac{1}{|\mathcal{X}_p|} \sum_{X_i \in \mathcal{X}_p} y_i \approx p. \tag{1}$$

This is the definition of a well-calibrated classifier, which can be probabilistically defined as -

$$P(Y = 1 \mid g(X) = p) = p \quad \forall p \in [0, 1], \tag{2}$$

In practice, classifiers are often not perfectly calibrated. To address this, we introduce a calibration function $c$. A calibration function is a mapping $c : [0, 1] \to [0, 1]$, which is applied to the output of $g$ to yield a calibrated probability estimate.

The classifier $g$ can be calibrated using a post-hoc calibration function $c$ such that

$$P(Y = 1 \mid c(g(X)) = \theta) = \theta \quad \forall \theta \in [0, 1], \tag{3}$$

where $\theta$ represents the calibrated probability estimate produced by applying $c$ to the original prediction $g(x)$.

The objective of post-hoc calibration methods is to find a calibration function $c$ that satisfies Eq. 3 as closely as possible.

### 3.2 Reliability Diagrams

A reliability diagram is a graphical representation used to assess the calibration of probabilistic classifiers on a dataset. It plots the empirical probability from the dataset $\mathcal{D} = \{(X_i, y_i) \mid i = 1, 2, \ldots, n\}$ against the predicted probability $g(X_i)$ for a classifier $g$, providing a visual indication of how well the classifier's predicted probabilities align with the empirical likelihood of $y_i = 1$ in $\mathcal{D}$.

Let's consider $p_i$ to be the predicted probability of $y_i = 1$ for some $X_i$. By definition, $p_i = g(X_i)$. To construct a reliability diagram, the set of predicted probabilities $P = \{p_1, p_2, \ldots, p_n\}$ corresponding to $(X_1, X_2, \ldots, X_n)$ and the true labels $\{y_1, y_2, \ldots, y_n\}$ are used. $P$ is first divided into $M$ equally spaced bins. Here, $M$ is a parameter of the reliability diagram and can be chosen based on the number of samples in the dataset. $B_m$ denotes the set of indices of samples whose predicted probabilities fall into the $m$-th bin.

$$B_m = \left\{ i \mid p_i \in \left( \frac{n-1}{M}, \frac{n}{M} \right] \right\} \tag{4}$$

For each bin $B_m$, the empirical probability is computed, which is the proportion of samples within the bin for which the true label $y_i = 1$. This is given by:

$$\mu_y(B_m) = \frac{1}{|B_m|} \sum_{i \in B_m} y_i \tag{5}$$

The average predicted probability, $\mu_p(B_m)$ is also computed for each bin $B_m$:

$$\mu_p(B_m) = \frac{1}{|B_m|} \sum_{i \in B_m} p_i \tag{6}$$

The set of points $(\mu_p(B_m), \mu_y(B_m))$ for each bin $m$ is then plotted on a graph and represents the reliability diagram for the given classifier $g$ and the dataset $\mathcal{D}$.

A perfectly calibrated model would result in points lying on the line $y = x$. The sample reliability diagram is illustrated in Fig. 1

### 3.3 Expected Calibration Error (ECE)

The Expected Calibration Error (ECE) is a scalar summary of the reliability diagram[15]. It quantifies the calibration performance by the average absolute difference between the predicted probabilities and the empirical probabilities across all bins of the reliability diagram.

$$\text{ECE} = \sum_{m=1}^{M} \frac{|B_m|}{M} \left| \mu_y(B_m) - \mu_p(B_m) \right| \tag{7}$$

The ECE provides a single number that summarizes the overall calibration error, with lower values indicating better calibration.

## 4 METHOD

In this section, we introduce ForeCal, a novel post-hoc calibration algorithm. The ForeCal algorithm addresses the intrinsic miscalibration issues in DNNs by applying a post-hoc transformation function $c_{RF}$ to the uncalibrated probabilities output by the DNN. The key to ForeCal lies in treating calibration as a regression problem and leveraging the unique properties of Random Forests to learn a monotonic calibration function $c_{RF}$ that improves calibration while preserving the discriminative power of the base model.

Consider calibration training dataset $\mathcal{D}_C = \{(X_i, y_i) \mid i = 1, 2, \ldots, n\}$, where $X_i$ is the input feature vector and $y_i$ is the binary outcome. We also have the predicted probabilities $\{p_1, p_2, \ldots, p_n\}$ corresponding to the input feature vectors $\{X_1, X_2, \ldots, X_n\}$ from the base model, $g$ where $p_i = g(X_i)$. As done while constructing the reliability diagram, the set of predicted probabilities $\{p_1, p_2, \ldots, p_n\}$ (corresponding to $(X_1, X_2, \ldots, X_n)$ and the true labels $\{y_1, y_2, \ldots, y_n\}$) is first divided into $M$ equally spaced bins as in Equation 4 for the construction of the reliability diagram.

For each bin $B_i$, $P$ bootstrap samples are then generated by sampling with replacement. For each bootstrap sample $S_{i,k}$, the mean predicted probability and the empirical probability are calculated as in Equation 8. This is similar to the calculation of the empirical probability and the mean predicted probability in the reliability diagram construction in Equations 5 and 6. However, here, the calculation is done for each bootstrap sample $S_{i,k}$ instead of the entire bin $B_i$ at once - yielding $P$ estimates of the mean predicted probability and the empirical probability for each bin $B_i$.

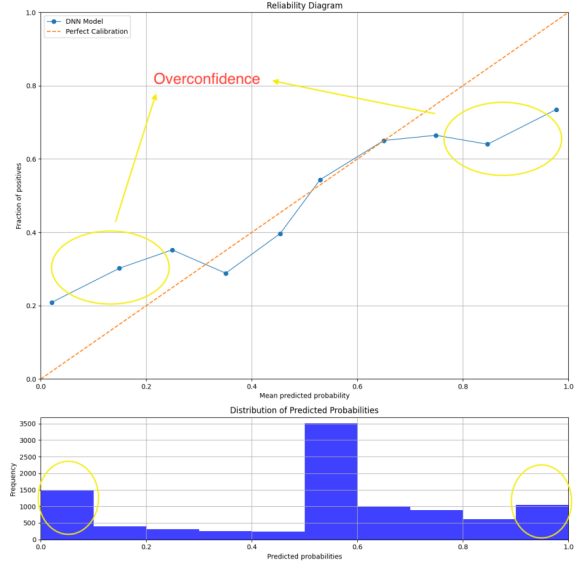

**Figure 1:** Reliability diagram showing the calibration of a DNN classifier trained on the Adult dataset from the UCI ML repository[2]. The predicted probabilities are divided into 10 bins, and the empirical probabilities are plotted against the average predicted probabilities for each bin. The diagonal line represents perfect calibration. In regions where the predicted and empirical probabilities are lower, the classifier skews the predicted probabilities towards 0. This results in the empirical probabilities being higher than the predicted probabilities in these regions. On the other hand, in regions where the predicted probabilities are higher, the classifier skews the predicted probabilities towards 1, resulting in the empirical probability being lower than the predicted probabilities.

$$\mu_p(S_{i,k}) = \frac{1}{|S_{i,k}|} \sum_{a \in S_{i,k}} p_a$$
$$\mu_y(S_{i,k}) = \frac{1}{|S_{i,k}|} \sum_{a \in S_{i,k}} y_a \tag{8}$$

The mean predicted probability and the empirical probability for each bootstrap sample $S_{i,k}$ for each bin $B_i$ are then aggregated to form the regression dataset $\mathcal{T}$.

$$\mathcal{T} = \bigcup_{i=1}^{N} \bigcup_{k=1}^{P} \left\{ (\mu_p(S_{i,k}), \mu_y(S_{i,k})) \right\} \tag{9}$$

This regression training dataset $\mathcal{T}$ is then used to train a Random Forest Regression model with $\mu_p(S_{i,k})$ as the independent variable and $\mu_y(S_{i,k})$ as the dependent or target variable while enforcing a monotonicity constraint between the two. The samples in the training set are weighted proportional to the number of samples in the bin they were sampled from $|B_i|$. Since we are splitting the predicted probabilities into equally spaced bins, the number of samples in each bin can be different. This sample weighting during Random forest training ensures that bins with fewer samples do not have an undue influence on the calibration function. It also makes the calibration function less sensitive to the number of bins. This training process yields $c_{RF}$ - a calibration function that transforms uncalibrated probability outputs into calibrated probabilities.

The bootstrapping step is unique to ForeCal as compared to other binning methods like histogram binning or BBQ. It allows ForeCal to capture the uncertainty in the calibration training dataset $\mathcal{D}_C$ itself and generate a more robust calibration function. Histogram binning or BBQ based calibration functions are highly sensitive to the bin thresholds and the dataset $\mathcal{D}_C$ itself[10].

Enforcing the monotonicity constraint during the training of the Random Forest Regression model ensures that $c_{RF}$ is a non-decreasing function of the uncalibrated probabilities. This is not a sufficient condition for preserving the ranks of the uncalibrated probabilities. However, given that Random Forest Regression is based on decision trees, which are piecewise constant functions, enforcing a strictly increasing monotonic constraint is not possible[4]. However, with enough trees in Random Forests, the piecewise constant function can approximate a strictly increasing. This preserves the discriminative power of the base model. The complete procedure for dataset construction is outlined in Algorithm 1.

The ForeCal algorithm exploits two key properties of Random Forest Regression to ensure that the calibration function $c_{RF}$ satisfies the desirable properties of a post-hoc calibration function.

The first is that a Random Forest Regression model exhibits *range preservation*. This property ensures that the predicted values from the trained model are bounded within the range of the target variable in the training data. In many applications, this lack of extrapolation is considered a drawback; however, for ForeCal, this is crucial since we want the output of $c_{RF}$ to be bounded in $[0, 1]$ to be interpreted as probabilities. Other regression methods like boosting or neural networks can predict values outside the training range, leading to outputs that cannot be interpreted as probabilities. To solve this problem they require additional post-processing steps like the softmax transformation which might distort calibration. Random Forests, on the other hand, do not require such post-processing steps to ensure bounded outputs and are unique among modern machine learning methods in this regard.

The second property is that enforcing monotonic constraints on decision tree splits in Random Forest during training ensures that for any $x_1$ and $x_2$ such that $x_1 \leq x_2$, we have $c_{rf}(x_1) \leq c_{rf}(x_2)$. Proofs for both these properties are provided in the Appendix.

[? ]

## 5 EXPERIMENTS AND RESULTS

### 5.1 Experiment Setup

To validate the effectiveness of ForeCal, we conducted experiments on 43 tabular classification datasets from the UCI ML repository [12]. The datasets were chosen on the basis of the classification and tabular nature and the availability of at least 1000 instances. The datasets were from diverse domains such as finance, healthcare, and marketing. In case the target was not binary, we first converted it to binary by considering the most frequent class as the positive

---

**Algorithm 1** ForeCal Algorithm

0: **Input:** Predicted probabilities $\{p_1, p_2, \ldots, p_n\}$, Corresponding outcomes $\{y_1, y_2, \ldots, y_n\}$, Number of bins $N$, Number of bootstrap samples per bin $P$
0: **Output:** Calibrated random forest model $c_{RF}$
0: **procedure** CREATEDATASET($\{p_i\}, \{y_i\}, N, P$)
0:     Initialize dataset $\mathcal{T} \leftarrow \emptyset$
0:     Divide $\{p_i\}$ into $N$ bins with equal width
0:     **for** $i = 1$ to $N$ **do**
0:         Define bin $B_i = \left\{ j \mid p_j \in \left[ \frac{i-1}{N}, \frac{i}{N} \right) \right\}$
0:         **for** $k = 1$ to $P$ **do**
0:             Sample with replacement $(p_j, y_j)$ $|B_i|$ times from $B_i$ to form $S_{i,k}$
0:             $\mu_p(S_{i,k}) = \frac{1}{|S_{i,k}|} \sum_{j \in S_{i,k}} p_j$
0:             $\mu_y(S_{i,k}) = \frac{1}{|S_{i,k}|} \sum_{j \in S_{i,k}} y_j$
0:             Append $(\text{mean}_p(S_{i,k}), \text{mean}_y(S_{i,k}))$ to $\mathcal{T}$
0:         **end for**
0:     **end for**
0:     **return** $\mathcal{T}$
0: **end procedure**
0: **procedure** TRAINRANDOMFOREST($\mathcal{T}$)
0:     Train Random Forest with monotonicity constraint on $\mathcal{T}$ with $\text{mean}_p$ as independent variable and $\text{mean}_y$ as dependent variable
0:     **return** $c_{RF}$
0: **end procedure**=0

---

class. The datasets were preprocessed and split into training(50%), calibration(30%) and testing(20%) sets. For each dataset, we trained a DNN classifier with 2 hidden layers of 100 neurons each using the training data. This served as our base model. We then calculated the predicted probabilities of the base model on the calibration set and trained a post-hoc calibration model using the ForeCal algorithm. We also trained calibration models using Isotonic regression, Platt scaling, BBQ, Histogram Binning, and Temperature scaling for comparison.

| Metric | Min | 25th Percentile | Median | 75th Percentile | Max |
|---|---|---|---|---|---|
| Base model AUC | 0.61 | 0.79 | 0.91 | 0.98 | 1.00 |
| Base model ECE | 0.00 | 0.01 | 0.06 | 0.10 | 0.30 |
| Dataset size | 1013 | 3170 | 8760 | 24510 | 299285 |

**Table 1: Summary statistics of the datasets used.**

### 5.2 Results and Discussion

Table 2 summarizes the median reduction in ECE and AUC over the base model for each calibration method across all 43 datasets. Forecal outperforms other calibration methods in terms of median ECE improvement over the base model except for Histogram binning. Histogram binning is surprisingly more effective in reducing ECE, although it comes at a significant cost of 4.12% reduction in AUC. This might not be acceptable in most real-world scenarios. Among methods that have a reasonably low reduction in AUC, Forecal is the best-performing method.

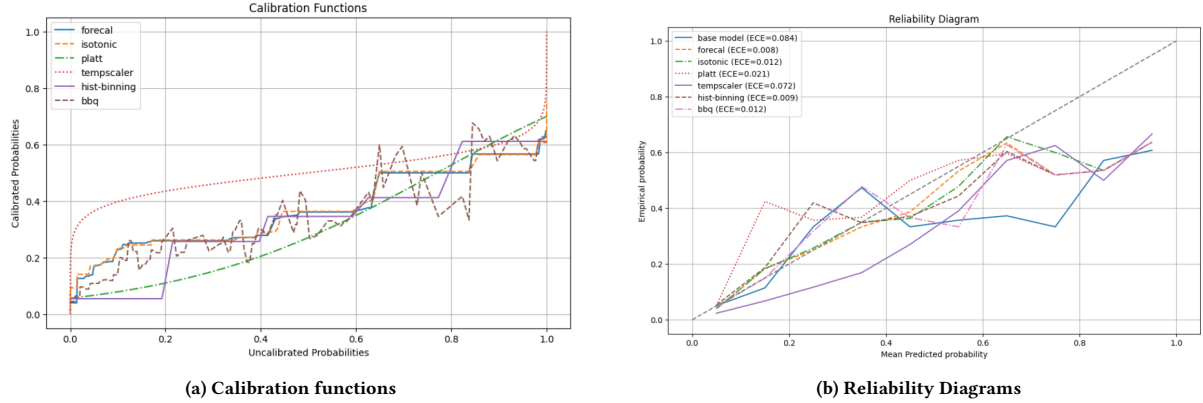

(a) Calibration functions

(b) Reliability Diagrams

**Figure 2:** Results on the Bank Marketing [13] dataset from the UCI ML repository. It is evident that although Forecal and IsoReg have better ECE values, their AUC has slightly deteriorated from the base model. The reason is evident in the calibration curves in Figure 2a. Since both these methods approximate the calibration function using a piecewise constant function, the learned calibration function is not strictly increasing w.r.t. the uncalibrated probabilities. This implies that for some cases, an increase in the uncalibrated probability does not result in an increase in the calibrated probability. This compromises the relative ranking of the uncalibrated probabilities and hence the AUC.

Both Forecal and Isotonic regression approximate the calibration function using a piecewise constant function and hence only guarantee weak monotonicity. Hence, they suffer slightly in terms of AUC. Platt scaling and Temperature scaling on the other hand do not have this slight deterioration in AUC as they assume a linear calibration function that enforces strict monotonicity. Histogram binning and BBQ, on the other hand, do not guarantee even weak monotonicity and hence have a significant reduction in AUC. A visual intuition for the shape of the calibration functions learnt by different methods is shown in Figure 2a.

Additionally, Forecal has the lowest standard error in ECE reduction across all datasets. This indicates that the performance of Forecal is more stable across different datasets compared to other methods.

| Method | Median Change in ECE(%) | Median change in AUC(%) |
|---|---|---|
| bbq | -69.60 ± 0.20 | -2.08 ± 0.02 |
| forecal | -75.93 ± 0.18 | -0.65 ± 0.01 |
| hist-binning | -77.26 ± 0.24 | -4.12 ± 0.05 |
| isotonic | -74.49 ± 0.19 | -0.07 ± 0.00 |
| platt | -60.85 ± 0.26 | -0.00 ± 0.00 |
| tempscaler | -28.26 ± 0.38 | 0.00 ± 0.00 |

**Table 2:** Median change in percentage points across all 43 datasets in ECE and AUC over the base model for each calibration method. The experiment was run 50 times with different random seeds to estimate metrics.

## 6  CONCLUSION

This paper introduced ForeCal, a novel post-hoc calibration model that leverages the power of Random Forests to address the miscalibration issues commonly observed in Deep Neural Network (DNN) models. ForeCal was rigorously tested on diverse datasets and demonstrated significant reductions in Expected Calibration Error(ECE) over traditional approaches such as Isotonic regression, Platt scaling, and Temperature scaling.

We also benchmark the rank preservation property of ForeCal against other calibration methods by measuring the change in AUC between the base model and the calibrated model. We find that methods like Isotonic regression and Forecal which approximate the calibration function using a piecewise constant function do not strictly preserve the ranks of the uncalibrated probabilities and hence might compromise the AUC and accuracy of the base model slightly. However, in our experiments, this trade-off was acceptable given the significant improvement in calibration. We also show that methods with stronger monotonicity constraints outperform those with weaker or no monotonicity constraints in terms of AUC preservation.

In the current work, we have not explored hyperparameter tuning for the Random Forest model used in ForeCal for each dataset. We have also not touched upon problems where incorporating extra features apart from the predicted probabilities from the base model might help in learning a better calibration function. These are interesting directions for future work and will further enhance the performance of ForeCal.

## A  APPENDICES

**Proposition 1:** Let $X$ denote the feature matrix and $y$ denote the target variable in the training dataset, where $y = \{y_1, y_2, \ldots, y_n\}$. Let $\hat{y}$ be the predicted value of the target variable for a given input $x$ using a random forest regression model trained on $X$ and $y$. The range preservation property can be formally stated as follows:

$$\min(y) \leq \hat{y} \leq \max(y) \quad \forall x$$

**Proof:**

1. **Individual Tree Predictions:**

$$\forall \text{ leaf node } l, \quad \hat{y}_l = \frac{1}{|L_l|} \sum_{j \in L_l} y_j, \quad \Rightarrow \min(y) \leq \hat{y}_l \leq \max(y) \tag{10}$$

where $L_l$ is the set of indices of samples in leaf node $l$.

**2. Individual Tree Predictions:**

$$\forall \text{ tree } t, \quad \hat{y}_t(\mathbf{x}) = \sum_{l \in \text{ leaves}} I(\mathbf{x} \in l) \hat{y}_l \Rightarrow \min(\mathbf{y}) \leq \hat{y}_t(\mathbf{x}) \leq \max(\mathbf{y}) \tag{11}$$

**3. Random Forest Prediction:**

$$\hat{y}(\mathbf{x}) = \frac{1}{M} \sum_{t=1}^{M} \hat{y}_t(\mathbf{x}) \Rightarrow \min(\mathbf{y}) \leq \hat{y}(\mathbf{x}) \leq \max(\mathbf{y}) \tag{12}$$

Thus, the prediction $\hat{y}(\mathbf{x})$ of the random forest regression model is bounded within the range of the target variable in the training dataset.

$$\boxed{\min(\mathbf{y}) \leq \hat{y}(\mathbf{x}) \leq \max(\mathbf{y}) \quad \forall \mathbf{x}}$$

**Proposition 2:** *Enforcing monotonic splits within trees ensures weak monotonicity of the calibration function.*

**Proof:**

Consider a decision tree $T$ with input feature $x$ and output $y$, constrained to be monotonically increasing. Let a split at $x = s$ partition the data into $L_s$ and $R_s$, such that:

$$\forall x_1, x_2 \in L_s, \, x_1 < x_2 \Rightarrow y_1 \leq y_2,$$
$$\forall x_1, x_2 \in R_s, \, x_1 < x_2 \Rightarrow y_1 \leq y_2, \tag{13}$$
$$\forall x_L \in L_s, \, x_R \in R_s, \, x_L < x_R \Rightarrow y_L \leq y_R.$$

For a tree with $n$ splits maintaining monotonicity, an $(n+1)$-th split that preserves the monotonic constraint refines $L_n$ or $R_n$ without disrupting the overall monotonicity.

By induction, if all splits are monotonic, the entire tree $T$ is monotonically increasing.

Since a Random Forest model is the average of such monotonic trees, and the sum of monotonic functions is monotonic, the final prediction of the Random Forest is also monotonic.